\documentclass{article} 
\usepackage{iclr2027_conference, times}

\usepackage{amsmath,amsfonts,bm}

\def\eqref#1{equation~\ref{#1}}

\def\1{\bm{1}}

\DeclareMathAlphabet{\mathsfit}{\encodingdefault}{\sfdefault}{m}{sl}
\SetMathAlphabet{\mathsfit}{bold}{\encodingdefault}{\sfdefault}{bx}{n}

\usepackage{hyperref}
\usepackage{url}
\usepackage{amsmath,amssymb}
\usepackage{float}
\usepackage{graphicx,multirow,xcolor,colortbl}
\usepackage{wrapfig}
\definecolor{PVYellow}{HTML}{FFF7E6}
\definecolor{GainGreen}{HTML}{278C68}
\newcommand{\pv}[2]{\textbf{#1}\,\raisebox{-0.3ex}{\scalebox{0.70}{\textcolor{GainGreen}{+#2}}}}
\definecolor{LossRed}{HTML}{B23A48}
\newcommand{\pvloss}[2]{\textbf{#1}\,\raisebox{-0.3ex}{\scalebox{0.70}{\textcolor{LossRed}{-#2}}}}

\title{Geometry-Supervised Visual Representation Learning for Multi-Phenotype Lesion Interpretation in Medical VLMs}

\author{%
\parbox[t]{\dimexpr\textwidth-2\tabcolsep\relax}{%
\centering
\normalfont
\setlength{\tabcolsep}{10pt}
\begin{tabular}[t]{@{}cccc@{}}
\textbf{Hao Wang}$^{1,2}$\thanks{These authors contributed equally to this work and share first authorship.} &
\textbf{Qiwei Zeng}$^{2}$\footnotemark[1] &
\textbf{Shuchang Ye}$^{1}$ &
\textbf{Jinghao Lin}$^{3}$ \\[4pt]
\textbf{Yuezhe Yang}$^{2}$ &
\textbf{Yige Peng}$^{2}$ &
\textbf{Jinman Kim}$^{1}$\thanks{Corresponding authors: Jinman Kim and Lei Bi.} &
\textbf{Lei Bi}$^{2}$\footnotemark[2] \\[7pt]
\multicolumn{4}{c}{$^{1}$The University of Sydney} \\
\multicolumn{4}{c}{$^{2}$Shanghai Jiao Tong University} \\
\multicolumn{4}{c}{$^{3}$Northeastern University}
\end{tabular}%
}%
}

\iclrfinalcopy 
\begin{document}

\maketitle

\begin{abstract}
Medical vision-language models (VLMs) have shown increasing potential for clinical image interpretation. 
However, these models still struggle to interpret multi-phenotype lesions whose diagnosis requires the joint assessment of multiple pathological phenotypes. 
Existing vision–language alignment methods produce visual representations that fail to preserve anatomical hierarchies and relationships among phenotypic subclasses. 
This stems from their reliance on semantic supervision, which lacks geometric constraints to preserve these relationships in the visual embedding space.
Moreover, the sparsity of lesion-related anatomical and phenotypic representations makes it difficult for medical VLMs to capture important diagnostic evidence.
To address these limitations, we propose \textbf{PureVision}, a geometry-supervised visual representation learning framework for multi-phenotype lesion interpretation in medical VLMs. 
It combines a geometry-supervised representation learning module, \textbf{PureEyes}, and an anatomy-guided evidence aggregation module, \textbf{PureNeurons}.
PureEyes provides geometric supervision through ideal visual semantic distribution that encode anatomical hierarchies and phenotypic subclass relationships.
PureNeurons projects visual representations into the learned latent space, using their positions to selectively aggregate lesion-specific anatomical and phenotypic evidence.
Experiments on \textit{LIDC-IDRI}, \textit{CBIS-DDSM}, and \textit{3DReasonKnee} demonstrate that PureVision improves lesion grounding and phenotype characterization in visual question answering and radiology report generation.
Code is available at: https://anonymous.4open.science/r/purevision-06C2.
\end{abstract}

\section{Introduction}

Medical vision-language models (VLMs) have been developed to assist clinicians in diagnosing diseases based on medical images \cite{wu2025towards}. 
Through large-scale pretraining on medical image–diagnosis pairs, they have learned to diagnose diseases characterized by a single pathological phenotype visible in medical images \cite{li2023llava,sellergren2025medgemma}. 
However, existing medical VLMs exhibit inaccurate lesion grounding and confusion in pathological phenotype descriptions when dealing with complex diseases whose diagnosis requires simultaneous consideration of anatomical structures and multiple pathological phenotypes. 
We attribute these limitations to the lack of fine-grained visual representations that can effectively distinguish anatomical structures and diverse pathological phenotypes.

To better capture anatomical structures and pathological phenotypes, existing research has explored semantically guided visual representation learning \cite{xing2025mca,gu2025radalign}.
For anatomical representations, BiRD \cite{huang2024refer} uses segmentation-derived region annotations for referring and grounding instruction tuning, linking anatomical descriptions to localized image regions.
VividMed \cite{luo2025vividmed} uses phrase embeddings to prompt a localization module that predicts segmentation masks and bounding boxes, explicitly grounding anatomical concepts in images.
For pathological phenotypes, MCA-RG \cite{xing2025mca} aligns visual features with pathology concepts and introduces a pathology–anatomy matching loss to emphasize disease-related regions.
RadAlign \cite{gu2025radalign} uses learnable concept tokens to extract attribute-specific visual features and aligns them with textual descriptions of diagnostic criteria through contrastive learning.

However, the distribution of these text embeddings in the semantic representation space deviates from the ideal distribution of visual representations determined by anatomical structures and pathological phenotypes in the vision embedding space. 
We assume that under an ideal distribution, the geometric relation of representations should align with the spatial correlation of anatomical structures and pathological phenotypes: 
the close geometric relationship of anatomical structures is reflected as a short distance in the embedding space; 
the relationship between anatomical structures and their sub-structures should form a hierarchical representation; 
pathological phenotypes across different severity levels should form a smooth and continuous distribution in the embedding space, reflecting the gradual variation of their visual characteristics.
Text embeddings used for supervision lack these geometric attributes, leading to insufficient anatomical and phenotypic information in visual representations. 
Moreover, pathological regions are typically sparse and account for only a small proportion of all image patches. The pathological information represented by these sparse patches can be interfered with by the large number of normal patches, making it difficult to accurately localize pathological regions and distinguish their phenotypic characteristics.

\begin{figure}
    \centering
    \includegraphics[width=\linewidth]{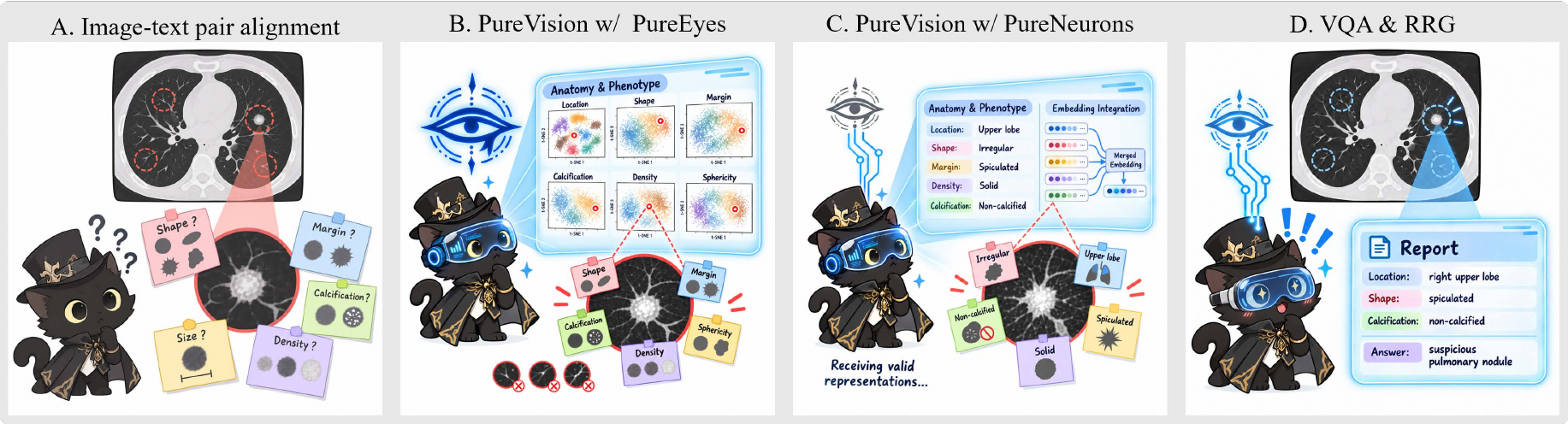}
    \vspace{-15pt}
    \caption{PureEyes structures the visual latent representation space, while PureNeurons captures and delivers lesion-relevant evidence.}
    \label{fig:overview}
    \vspace{-15pt}
\end{figure}

To bridge these gaps, we propose a representation learning framework for medical VLMs' visual encoders named \textbf{PureVision} as shown in Figure~\ref{fig:overview}, which uses ideal visual semantic distribution to constrain the visual embedding space and dynamically translate these anatomical and phenotypic representations to semantic space. 
The ideal visual semantic distribution reflects the actual relationships among anatomical structures and phenotypic subclasses. 
Firstly, to address the distribution deviation, we develop \textbf{PureEyes}, which directly supervises visual encoders using ideal visual semantic distribution that preserve anatomical hierarchies and relationships among phenotypic subclasses. 
Furthermore, to effectively capture and deliver sparse lesion information, we develop \textbf{PureNeurons}, which dynamically selects lesion-relevant patches and delivers their anatomical and phenotypic evidence through semantic fusion.
Experiments are conducted on \textit{LIDC-IDRI}, \textit{CBIS-DDSM}, and \textit{3DReasonKnee}, covering CT, mammography, and MRI of different lesion types. 
PureVision improves lesion grounding and phenotype characterization across visual question answering (VQA) and radiology report generation (RRG).
Our contributions are summarized as follows:

\noindent \textbf{(1)} We propose \textbf{PureVision}, a representation learning framework for medical VLM visual encoders that combines supervision based on ideal visual semantic distribution with selective information delivery for lesion grounding and pathological phenotype characterization.

\noindent \textbf{(2)} We develop \textbf{PureEyes}, a visual representation learning module that directly supervises separate anatomical and phenotypic encoders using ideal distributions to preserve anatomical hierarchies and continuity across ordered phenotypic subclasses.

\noindent \textbf{(3)} We introduce \textbf{PureNeurons}, an anatomy-guided information delivery module that selectively aggregates lesion-specific anatomical and phenotypic evidence to reduce interference from normal-tissue patches and delivers this evidence as soft semantic tokens to maintain the alignment with semantic embeddings.

\vspace{-10pt}
\section{Related Work}

\subsection{Medical VLMs in Radiology}

Medical VLMs adapt pretrained visual and language models to medical interpretation through domain-specific multimodal training. LLaVA-Med combines biomedical figure–caption alignment with instruction tuning to support medical visual conversation \cite{li2023llava}. LLaVA-Rad connects pretrained image and language components through a lightweight adapter and specializes the model using radiological image–text pairs \cite{zambrano2025clinically}. MedGemma incorporates medical image encoder adaptation into a Gemma-based multimodal model \cite{sellergren2025medgemma}. Complementing model adaptation, MedRAX integrates chest X-ray analysis tools with multimodal models for complex clinical querying \cite{fallahpour2025medrax}.

The scope of radiology VLMs also extends to volumetric inputs and anatomically structured interpretation. M3D-LaMed unifies multiple 3D tasks, including report generation, question answering, positioning, and segmentation, within a multimodal instruction-learning framework \cite{bai2024m3d}. Argus studies how encoder pretraining, visual token compression, and model and data scaling affect high-resolution CT report generation \cite{liu2025argus}. AOR introduces anatomy-centric, region-level reasoning to structure chest X-ray interpretation \cite{li2026aor}. Together, these studies address domain adaptation, volumetric processing, and reasoning organization. The following subsection examines the more specific mechanisms used to associate visual evidence with image regions and medical concepts.

\subsection{Fine-Grained Alignment in Medical VLMs}

Fine-grained alignment methods link local image regions to clinically meaningful descriptions. BiRD learns region-language correspondence from grounding instructions constructed from segmentation datasets, while MAIRA-2 associates radiological findings with bounding boxes during report generation \cite{huang2024refer,bannur2024maira}. Anatomy-aware methods further organize visual inputs around explicit anatomical regions. Reg2RG uses segmentation masks to extract region features for region-specific report generation, whereas MedRegion-CT combines region-guided tokenization with pseudo-mask supervision and structured lesion attributes \cite{chen2025large,kyung2025region}. Together, these approaches improve local correspondence through regional feature construction, explicit supervision, or region-specific inputs to the language model.

Concept-guided methods further associate visual representations with clinically meaningful semantics. MCA-RG introduces anatomical and pathological concept banks together with contrastive learning and feature gating, whereas RadAlign aligns visual concept tokens with textual diagnostic criteria to guide report generation \cite{xing2025mca,gu2025radalign}. Although these methods demonstrate the value of regional and concept-specific supervision, clinical text or concept embeddings do not necessarily impose an ideal structure on the visual latent space: anatomical hierarchies may remain poorly organized, and ordered phenotype subclasses may not exhibit the expected continuity. Moreover, lesion-specific evidence is often concentrated in only a few image patches and can therefore be diluted by normal-tissue representations. This combination of distribution mismatch and sparse lesion encoding limits the accurate interpretation of fine-grained pathological features after alignment.

\begin{figure}
    \centering
    \includegraphics[width=\linewidth]{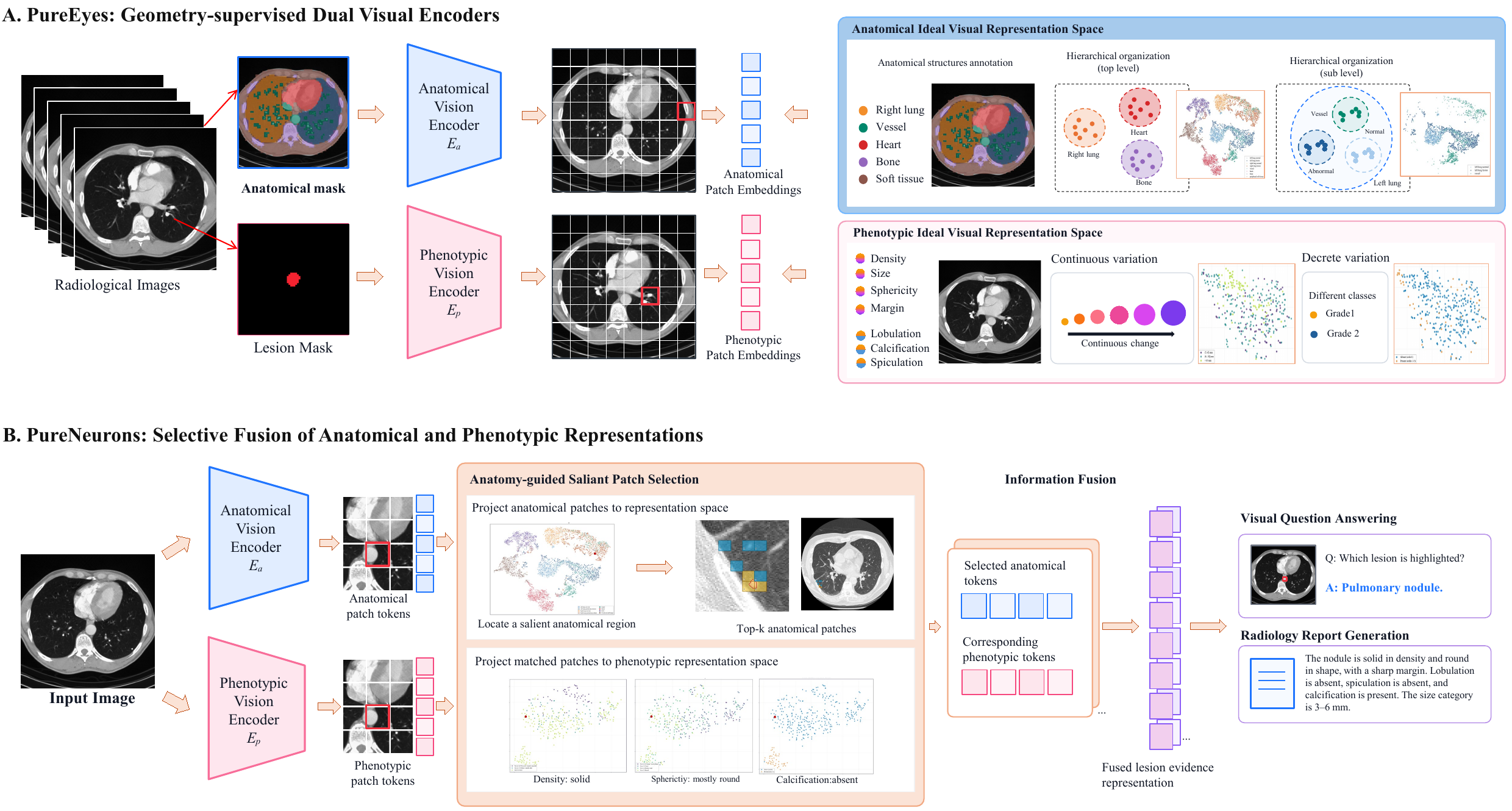}
    \vspace{-15pt}
    \caption{Overview of PureVision. (a) PureEyes shapes anatomical and phenotypic visual spaces with ideal geometries encoding anatomical hierarchies and phenotype relationships. (b) PureNeurons identifies lesion-relevant patches in these spaces and selectively fuses the corresponding anatomical and phenotypic evidence for downstream tasks.}
    \label{fig:flow}
    \vspace{-15pt}
\end{figure}

\section{Methodology}

\subsection{Overview}

PureVision consists of two components as shown in Figure~\ref{fig:flow}. \textbf{PureEyes} shapes anatomical and phenotypic visual representation spaces using annotation-defined ideal geometries, which maintain the relationship of anatomical structures and phenotypic subclasses. \textbf{PureNeurons} then identifies sparse lesion-relevant patches from the projection of the learned representation space and aligns their anatomical and phenotypic evidence into language-compatible soft tokens for downstream reasoning. The visual encoders are trained first and subsequently frozen during semantic alignment and inference.

\subsection{\textbf{PureEyes}: Building Ideal-geometry Supervised Visual Representation Space}

PureEyes learns complementary anatomical and phenotypic representations with visual encoders $E_A$ and $E_P$, preserving spatial correspondence while independently optimizing them toward different ideal visual geometries. The fine-tuning process consists of three stages: (1) lesion-centric phenotype geometry learning, where isolated lesion patches are organized according to phenotype-specific geometric constraints; (2) context-aware phenotype geometry learning, where the full image is encoded and the lesion-region representations are further constrained under the same ideal visual semantic distribution; and (3) anatomical geometry learning, where annotation-defined anatomical regions are organized according to their hierarchical relationships.

\paragraph{Phenotypic geometry supervision.}
Different phenotype types require different geometric structures. For continuous attributes, such as lesion size, normalized measurement differences define the target pairwise distances in the visual space:
\begin{equation}
\mathcal{L}_{\mathrm{cont}}^{g}
=
\mathbb{E}_{(n,m)}
\left[
\rho_{\delta}
\left(
d_{nm}^{g}
-
\alpha_g
\left|
\tilde{y}_{n}^{g}
-
\tilde{y}_{m}^{g}
\right|
\right)
\right],
\end{equation}
where $d_{nm}^{g}$ denotes the distance between normalized phenotype representations, $\tilde{y}_{n}^{g}$ is the normalized measurement, $\alpha_g$ controls the target distance scale, and $\rho_{\delta}$ is a robust penalty:
\begin{equation}
\rho_{\delta}(r)
=
\begin{cases}
\dfrac{r^2}{2\delta}, & |r| < \delta,\\[4pt]
|r|-\dfrac{\delta}{2}, & \text{otherwise}.
\end{cases}
\end{equation}
This objective matches visual-space distances to measurement-defined geometric distances, preserving continuous phenotypic variation.

For discrete phenotypes, representations of the same category are encouraged to form compact regions, while different categories are separated by a margin. For ordered phenotypes, an additional ranking constraint preserves the relative relationships among grades. Together, these objectives construct phenotype-specific geometries for continuous, ordinal, and categorical attributes; detailed formulations and implementation settings are provided in Appendix~\ref{app:implementation}.

We first apply these geometric constraints to representations extracted from isolated lesion patches to establish lesion-centric phenotype spaces. We then encode the complete image and apply the same constraints to lesion-overlapping patch representations, allowing phenotype geometry to incorporate surrounding anatomical context through self-attention. Lesion masks are used only to identify supervised patches during training.

\paragraph{Anatomical geometry supervision.}
The anatomical branch organizes tissues hierarchically, separating anatomical structures while preserving the relationship between normal and pathological states within the same anatomical parent. Its objective is
\begin{equation}
\mathcal{L}_A
=
\lambda_a\mathcal{L}_{\mathrm{anatomy}}
+\lambda_p\mathcal{L}_{\mathrm{parent}}
+\lambda_h\mathcal{L}_{\mathrm{hierarchy}}
+\lambda_d\mathcal{L}_{\mathrm{distill}}
+\lambda_c\mathcal{L}_{\mathrm{conf}}.
\end{equation}
Here, anatomy and parent supervision preserve tissue identity, hierarchical supervision separates normal and lesion states within the same parent region, distillation prevents unnecessary drift of non-lesion features, and confounding-tissue contrast improves discrimination from visually similar normal structures. Anatomical annotations, pseudo-labels, and lesion masks are used only as training supervision.

\subsection{\textbf{PureNeurons}: Grounding Sparse Lesion Position and Align Phenotypes to Semantic Representation}

PureNeurons captures sparse lesion-specific information from the structured visual spaces learned by PureEyes and delivers it to the language model as dense semantic evidence. It first trains a shared alignment layer to map anatomical and phenotypic patch representations into a common semantic space while preserving their spatial correspondence. Based on the aligned representations, PureNeurons then identifies lesion-relevant regions from anatomical evidence and selectively aggregates the corresponding anatomical and phenotypic information into language-compatible soft semantic representations for the decoder.

\paragraph{Lesion Grounding in the Aligned Space.}
The alignment layer associates different regions of the anatomical representation space with their corresponding text labels. During inference, each anatomical patch is projected into the same aligned space, where its semantic identity can be inferred from the text label it matches best. To identify lesion-related patches, we compare their strongest match to lesion labels with their strongest match to other anatomical labels:
\begin{equation}
\rho_i
=
\max_{c\in\mathcal{C}_A^{\mathrm{lesion}}} S_{i,c}^{A}
-
\max_{c\in\mathcal{C}_A^{\mathrm{other}}} S_{i,c}^{A},
\end{equation}
where $S_{i,c}^{A}$ denotes the correspondence between patch $i$ and text label $c$. A larger $\rho_i$ indicates that the patch is more likely to belong to a lesion region than to normal anatomy.

\paragraph{Probability-weighted semantic fusion.}
Within each anatomical or phenotype group, patch-level similarities are spatially aggregated and normalized across candidate descriptions. Rather than making a hard subtype prediction, PureNeurons retains all candidates with different weights. Their language-model input embeddings are fused as
\begin{equation}
v_{g,\ell}
=
\sum_{c\in\mathcal{C}_g}
p_{g,c}\,\bar u_{g,c,\ell},
\end{equation}
where $p_{g,c}$ is the semantic weight of subtype $c$ and $\bar u_{g,c,\ell}$ is its length-aligned token embedding.

The resulting anatomical and phenotype soft tokens, together with the estimated lesion location, are inserted after the visual tokens and before the task instruction. The visual tokens are retained, allowing the frozen decoder to integrate explicit lesion-specific evidence with the complete image context for downstream tasks.

\begin{table*}[t]
  \centering
  \tiny
  \setlength{\tabcolsep}{1pt}
  \renewcommand{\arraystretch}{1.18}
  \resizebox{\textwidth}{!}{
  \begin{tabular}{@{\hspace{4pt}}l|c|cc|cc|cc@{\hspace{4pt}}}
    \hline\hline
    \multirow{2}{*}{\textbf{Models}} & \multirow{2}{*}{\textbf{Scale}} & \multicolumn{2}{c|}{\textbf{LIDC-IDRI}} & \multicolumn{2}{c|}{\textbf{CBIS-DDSM}} & \multicolumn{2}{c}{\textbf{3DReasonKnee}} \\ \cline{3-8}
    & & Grounding Acc. & Phenotypes Acc. & Grounding Acc. & Phenotypes Acc. & Grounding Acc. & Phenotypes Acc. \\ \hline
    RadFM & 14B & 2.00 & 5.00 & 2.00 & 17.00 & 5.00 & 3.00 \\
    \rowcolor{PVYellow} \quad \textbf{w/ PureVision} & $\sim$14B & \pv{11.00}{9.00} & \pv{23.00}{18.00} & \pv{23.00}{21.00} & \pv{24.00}{7.00} & \pv{21.00}{16.00} & \pv{24.00}{21.00} \\ \hline
    LLaVA-Med & 7B & 14.00 & 27.79 & 13.00 & 30.83 & 12.00 & 16.00 \\
    \rowcolor{PVYellow} \quad \textbf{w/ PureVision} & $\sim$7B & \pv{23.00}{9.00} & \pv{44.86}{17.07} & \pv{24.5}{11.50} & \pv{37.50}{6.67} & \pv{32.00}{20.00} & \pv{31.00}{15.00} \\ \hline
    Lingshu & 7B & 36.00 & 33.86 & 30.50 & 12.00 & 28.00 & 27.00 \\
    \rowcolor{PVYellow} \quad \textbf{w/ PureVision} & $\sim$7B & \pv{42.50}{6.50} & \pv{50.57}{16.71} & \pv{44.00}{13.50} & \pv{35.50}{23.50} & \pv{44.00}{16.00} & \pv{40.00}{13.00} \\ \hline
    Hulu-Med & 4B & 32.50 & 42.14 & 30.50 & 7.67 & 35.00 & 26.00 \\
    \rowcolor{PVYellow} \quad \textbf{w/ PureVision} & $\sim$4B & \pv{47.50}{15.00} & \pv{51.14}{9.00} & \pv{46.00}{15.50} & \pv{35.33}{27.66} & \pv{44.00}{9.00} & \pv{43.00}{17.00} \\ \hline
    MedGemma & 4B & 21.50 & 34.79 & 38.50 & 8.50 & 25.00 & 25.00 \\
    \rowcolor{PVYellow} \quad \textbf{w/ PureVision} & $\sim$4B & \pv{28.00}{6.50} & \pv{51.50}{16.71} & \pv{43.00}{4.50} & \pv{32.00}{23.50} & \pv{42.00}{17.00} & \pv{31.00}{6.00} \\ \hline
    MedGemma1.5 & 4B & 11.50 & 10.50 & 24.00 & 10.00 & 29.00 & 25.00 \\
    \quad w/ RadAlign & $\sim$4B & 23.00 & 32.93 & 37.50 & 31.17 & 36.00 & 34.00 \\
    \quad w/ MCA-RG & $\sim$4B & 23.50 & 29.29 & 32.50 & 19.67 & 30.00 & 29.00 \\
    \quad w/ Reg2RG & $\sim$4B & 11.50 & 23.79 & 32.00 & 17.17 & 33.00 & 23.00 \\
    \rowcolor{PVYellow} \quad \textbf{w/ PureVision} & $\sim$4B & \pv{45.50}{34.00} & \pv{49.93}{39.43} & \pv{41.50}{17.50} & \pv{48.67}{38.67} & \pv{43.00}{14.00} & \pv{50.00}{25.00} \\ \hline\hline
  \end{tabular}}
  \vspace{-10pt}
  \caption{Comparison with pretrained medical VLMs and alignment methods on the VQA task.}
  \label{tab:vqa_comparison}

  \vspace{5pt}

  \centering
  \tiny
  \setlength{\tabcolsep}{1pt}
  \renewcommand{\arraystretch}{1.18}
  \resizebox{\textwidth}{!}{
  \begin{tabular}{@{\hspace{4pt}}l|c|cc|cc|cc@{\hspace{4pt}}}
    \hline\hline
    \multirow{2}{*}{\textbf{Models}} & \multirow{2}{*}{\textbf{Scale}} & \multicolumn{2}{c|}{\textbf{LIDC-IDRI}} & \multicolumn{2}{c|}{\textbf{CBIS-DDSM}} & \multicolumn{2}{c}{\textbf{3DReasonKnee}} \\ \cline{3-8}
    & & Grounding Acc. & Phenotypes Acc. & Grounding Acc. & Phenotypes Acc. & Grounding Acc. & Phenotypes Acc. \\ \hline
    RadFM & 14B & 0.00 & 0.00 & 0.00 & 5.57 & 0.00 & 3.83 \\
    \rowcolor{PVYellow} \quad \textbf{w/ PureVision} & $\sim$14B & \pv{8.00}{8.00} & \pv{9.70}{9.70} & \pv{9.50}{9.50} & \pv{16.43}{10.86} & \pv{10.93}{10.93} & \pv{5.46}{1.63} \\ \hline
    LLaVA-Med & 7B & 5.00 & 13.00 & 0.00 & 11.41 & 4.92 & 8.20 \\
    \rowcolor{PVYellow} \quad \textbf{w/ PureVision} & $\sim$7B & \pv{28.00}{23.00} & \pv{33.10}{20.10} & \pv{12.00}{12.00} & \pv{27.33}{15.92} & \pv{13.66}{8.74} & \pv{9.59}{1.39} \\ \hline
    Lingshu & 7B & 12.50 & 39.50 & 7.00 & 7.28 & 23.50 & 21.80 \\
    \rowcolor{PVYellow} \quad \textbf{w/ PureVision} & $\sim$7B & \pv{40.00}{27.50} & \pv{47.74}{8.24} & \pv{20.00}{13.00} & \pv{12.81}{5.53} & \pv{41.53}{18.03} & \pv{34.43}{12.63} \\ \hline
    Hulu-Med & 4B & 25.50 & 37.19 & 12.00 & 17.04 & 24.59 & 29.02 \\
    \rowcolor{PVYellow} \quad \textbf{w/ PureVision} & $\sim$4B & \pv{57.50}{32.00} & \pv{50.18}{12.99} & \pv{24.50}{12.50} & \pv{22.18}{5.14} & \pv{50.27}{25.68} & \pv{45.05}{16.03} \\ \hline
    MedGemma & 4B & 15.50 & 16.49 & 10.50 & 16.51 & 32.24 & 25.00 \\
    \rowcolor{PVYellow} \quad \textbf{w/ PureVision} & $\sim$4B & \pv{45.00}{29.50} & \pv{45.79}{29.30} & \pv{14.50}{4.00} & \pv{27.24}{10.73} & \pv{48.09}{15.85} & \pv{34.03}{9.03} \\ \hline
    MedGemma1.5 & 4B & 13.50 & 19.38 & 10.00 & 12.19 & 36.61 & 25.00 \\
    \quad w/ RadAlign & $\sim$4B & 15.50 & 37.37 & 13.50 & 25.92 & 48.63 & 28.96 \\
    \quad w/ MCA-RG & $\sim$4B & 11.50 & 33.08 & 24.00 & 18.40 & 45.36 & 27.32 \\
    \quad w/ Reg2RG & $\sim$4B & 13.00 & 34.26 & 11.00 & 21.14 & 48.09 & 21.31 \\
    \rowcolor{PVYellow} \quad \textbf{w/ PureVision} & $\sim$4B & \pv{20.00}{6.50} & \pv{44.61}{25.23} & \pv{39.50}{29.50} & \pv{31.55}{19.36} & \pv{50.27}{13.66} & \pv{37.77}{12.77} \\ \hline\hline
  \end{tabular}}
  \vspace{-10pt}
  \caption{Comparison with pretrained medical VLMs and alignment methods on the RRG task.}
  \label{tab:rrg_comparison}
  \vspace{-15pt}
\end{table*}

\section{Experiments}

\subsection{Dataset and Evaluation}

We conduct training and evaluation on three public datasets, including \textbf{LIDC-IDRI} \cite{armato2011lung}, \textbf{CBIS-DDSM} \cite{lee2017curated}, and \textbf{3DReasonKnee} \cite{sambara20253dreasonknee}, which collectively cover diverse imaging modalities, anatomical structures, and disease types. LIDC-IDRI \cite{armato2011lung} is a CT dataset for pulmonary nodules, containing 12120 samples. It covers 6 anatomical structures and 7 nodule phenotypes. CBIS-DDSM \cite{lee2017curated} is a mammography dataset for breast lesions, containing 2543 test samples. It covers 2 anatomical structures and 3 lesion phenotypes. 3DReasonKnee \cite{sambara20253dreasonknee} is an annotated knee MRI dataset for musculoskeletal disease assessment based on the OAI dataset \cite{lester2012osteoarthritis}, containing 7846 test samples. We select the meniscus injury subset, which covers 5 anatomical structures and 1 lesion phenotype. Based on these datasets, we construct visual question answering (VQA) and radiology report generation (RRG) evaluation sets to assess our method’s ability in lesion localization and phenotype characterization, including 2,600 VQA pairs (1,600 for LIDC-IDRI, 800 for CBIS-DDSM, and 200 for 3DReasonKnee) and 583 RRG instances (200 for LIDC-IDRI, 200 for CBIS-DDSM, and 183 for 3DReasonKnee). Specifically, for lesion localization evaluation, we divide each image into a $4 \times 4$ grid, requiring the model to identify the grid cell containing the lesion. We use \textit{Grounding Accuracy} and \textit{Phenotype Characterization Accuracy} to evaluate the model’s ability in lesion localization and pathological attribute characterization, respectively. Detailed information on benchmark distributions, anatomical structures' annotation, phenotype categories, VQA/RRG construction, and evaluation protocols is provided in Appendix~\ref{app:data}.

\subsection{Experimental Setup}

\begin{figure}
    \centering
    \includegraphics[width=\linewidth]{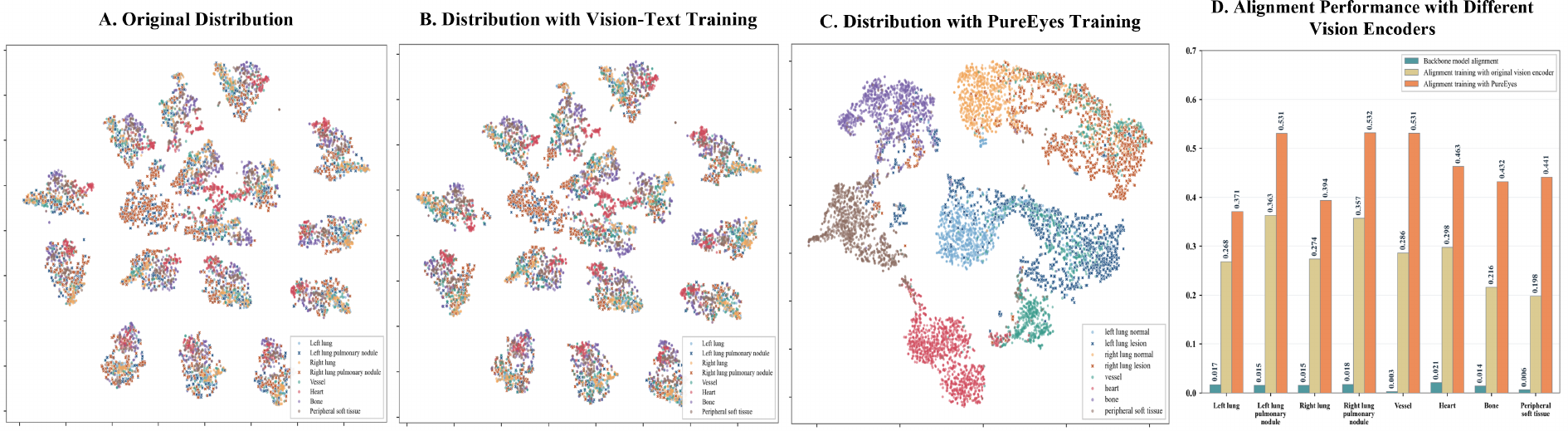}
    \vspace{-10pt}
    \caption{t-SNE visualization of anatomical representations and anatomy–text alignment across different vision encoders on LIDC-IDRI.}
    \label{fig:ana_tsne}
    \vspace{-10pt}
\end{figure}

\begin{figure}
    \centering
    \includegraphics[width=\linewidth]{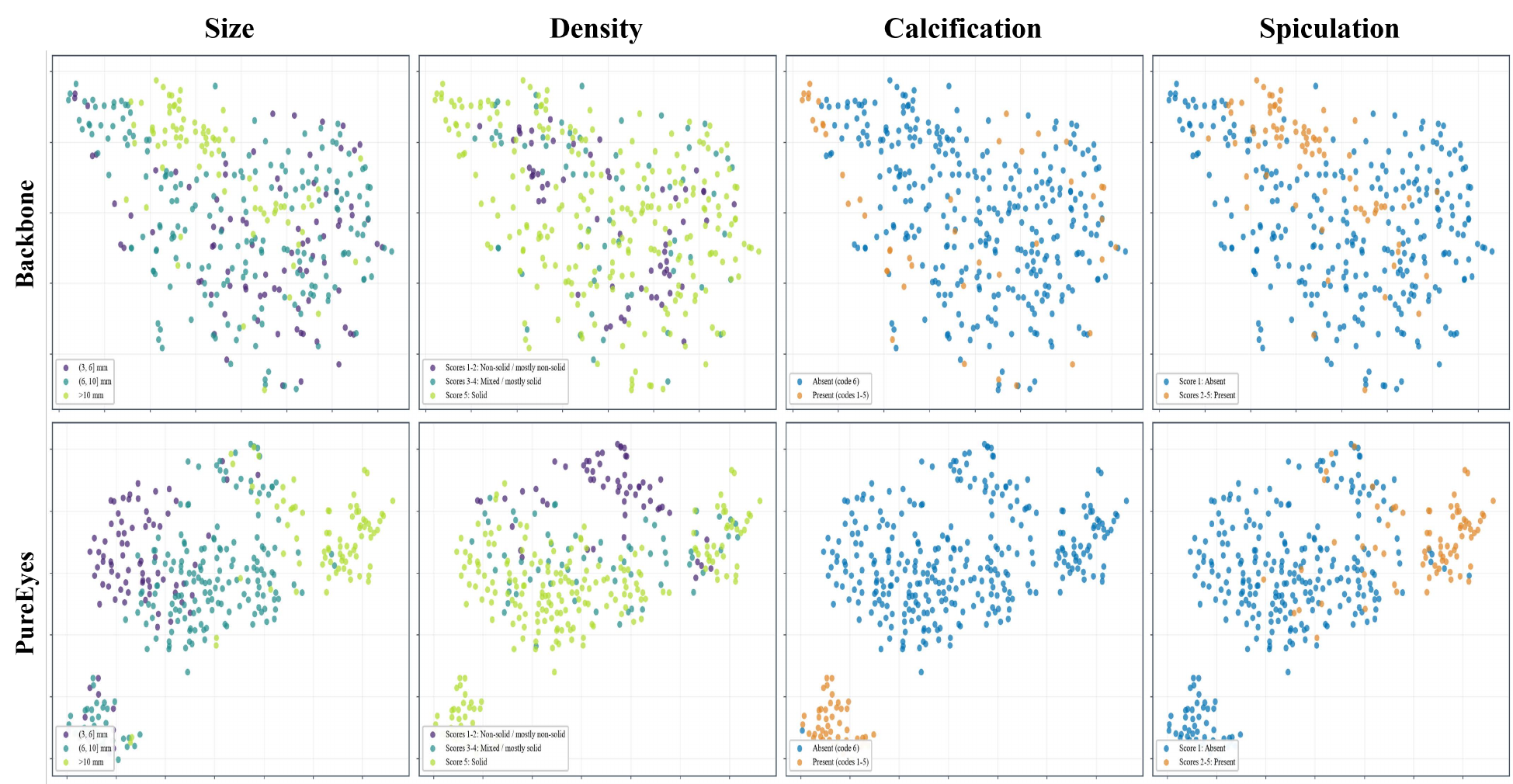}
    \vspace{-10pt}
    \caption{t-SNE visualization of pathological phenotype representations on LIDC-IDRI, comparing the backbone and PureEyes.}
    \label{fig:pheno_tsne}
\end{figure}

We conduct a comprehensive evaluation of PureVision from both downstream performance and module effectiveness. First, we compare PureVision with multiple pretrained medical VLMs \cite{wu2025towards,li2023llava,xu2026lingshu,jiang2025hulu,sellergren2025medgemma,sellergren2026medgemma} and existing representation alignment methods \cite{gu2025radalign,xing2025mca,chen2025large} on VQA and RRG tasks to evaluate lesion grounding and pathological phenotype characterization (Tables~\ref{tab:vqa_comparison} and~\ref{tab:rrg_comparison}). We then analyze PureEyes on LIDC-IDRI by visualizing anatomical and phenotypic embeddings with t-SNE and measuring anatomy--text similarity, assessing whether the learned representations preserve the intended anatomical structure and phenotype relationships (Figs.~\ref{fig:ana_tsne} and~\ref{fig:pheno_tsne}). For PureNeurons, we evaluate lesion patch recall and redundant patch rate under different selection budgets, together with ablations of lesion grounding and soft semantic fusion. Finally, we provide qualitative patch-selection examples and an end-to-end case study to illustrate how the learned representations and selected lesion evidence support downstream predictions. Implementation details are provided in Appendix~\ref{app:implementation}

\subsection{Comparison with Existing Methods and Alignment methods}

Table~\ref{tab:vqa_comparison} presents the comparison on the VQA task. To demonstrate the effectiveness and generalizability of PureVision across different medical VLMs, we apply PureVision to multiple pretrained medical VLM backbones. PureVision improves lesion grounding and pathological phenotype characterization across all evaluated backbones and datasets. For example, with MedGemma1.5, PureVision improves the grounding accuracy and phenotype characterization accuracy from 11.50/10.50 to 45.50/49.93 on LIDC-IDRI, from 24.00/10.00to 41.50/48.67 on CBIS-DDSM, and from 29.00/25.00 to 43.00/50.00 on 3DReasonKnee, respectively. Across all evaluated backbones and datasets, PureVision achieves average improvements of 14.19 percentage points (pp) and 18.94 pp in grounding accuracy and phenotype characterization accuracy, respectively, demonstrating its effectiveness in enhancing visual representations of pretrained medical VLMs. Furthermore, to compare PureVision with existing visual-text alignment approaches, we apply representative alignment methods, including RadAlign, MCA-RG, and Reg2RG, under the same MedGemma1.5 backbone. PureVision achieves the highest grounding and phenotype characterization accuracies across all three datasets, demonstrating that the proposed supervision based on the ideal visual semantic distribution and the extraction of sparse lesion-specific information can effectively enhance the model’s capability in lesion localization and description. Detailed phenotypes evaluation results are provided in Appendix~\ref{app:detailed_phenotypes_acc}.

Table~\ref{tab:rrg_comparison} further reports the comparison on the RRG task. To demonstrate the effectiveness of PureVision for open-ended radiology report generation, we apply PureVision to different pretrained medical VLM backbones. PureVision improves lesion grounding and pathological phenotype characterization across all evaluated backbones and datasets, with particularly pronounced gains in lesion localization. For example, with MedGemma1.5, PureVision improves the grounding and phenotype accuracies from 13.50/19.38 to 20.00/44.61 on LIDC-IDRI, from 10.00/12.19 to 39.50/31.55 on CBIS-DDSM, and from 36.61/25.00 to 50.27/37.77 on 3DReasonKnee. Similar improvements are observed across other pretrained backbones, demonstrating that the enhanced visual representations learned by PureVision can be effectively transferred to free-form radiology report generation. Furthermore, compared with existing representation alignment methods under the same MedGemma1.5 backbone, PureVision achieves the highest grounding and phenotype characterization accuracies across all three datasets.

\subsection{PureEyes Supervised Vision Encoder Performance}

Figure~\ref{fig:ana_tsne} compares anatomical patch representations on LIDC-IDRI. The original MedGemma1.5 encoder produces locally clustered but highly entangled representations, and vision--text alignment improves cross-modal similarity without substantially reorganizing this anatomical geometry. In contrast, PureEyes forms compact and well-separated anatomical clusters, with finer subregions for normal lung tissue, pulmonary lesions, and vessels. Normal and abnormal lung patches are also clearly separated (Appendix~\ref{app:appendix_pulmonary_tsne}), while the partial overlap between nodules and vessels reflects their similar appearance in individual CT slices. The rightmost panel further shows improved visual--text alignment across anatomical categories, with left- and right-lung nodule similarities increasing from 0.363/0.357 to 0.531/0.532. These results show that PureEyes learns a more structured anatomical space that better supports semantic alignment.

Figure~\ref{fig:pheno_tsne} compares phenotype representations produced by the original MedGemma1.5 vision encoder and the PureEyes-trained encoder. We visualize two continuously or ordinally varying phenotypes, size and density, and two discrete phenotypes, calcification and spiculation. With the original encoder, different size and density levels are scattered without clear ordering, while the discrete categories substantially overlap. In contrast, PureEyes arranges size and density representations along coherent progression trajectories, with adjacent levels remaining close, and organizes calcification and spiculation categories into more compact and distinguishable clusters. These results show that PureEyes captures both gradual phenotype transitions and discrete category boundaries. Visualizations of sphericity, margin, and lobulation are provided in Appendix~\ref{app:additional_phenotype_distribution}.

\begin{table*}[t]
    \centering
    \tiny
    \setlength{\tabcolsep}{2pt}
    \renewcommand{\arraystretch}{1.5}
    \resizebox{\textwidth}{!}{
    \begin{tabular}{@{\hspace{4pt}}c|cc|l|cc@{\hspace{4pt}}}
        \hline\hline
        \multicolumn{3}{c|}{\textbf{Lesion Extraction Efficiency}} & \multicolumn{3}{c}{\textbf{Lesion Information Delivery Ablation}} \\
        \cline{1-3} \cline{4-6}
        \textbf{Selected Patches ($K$)} & \textbf{Lesion Patch Recall} & \textbf{Redundant Patch Rate} & \textbf{Method} & \textbf{Grounding Acc.} & \textbf{Phenotypes Acc.} \\
        \hline
        4  & 60.11 & 45.23 & MedGemma1.5 & 11.50 & 10.50 \\
        \rowcolor{PVYellow} 8  & 67.64 & 65.81 & \quad w/o PureNeurons & 32.50 & 36.57 \\
        12 & 68.47 & 75.82 & \quad w/ lesion grounding & 40.00 & 38.43 \\
        \rowcolor{PVYellow} 16 & 68.84 & 81.23 & \quad w/ lesion grounding and soft semantic fusion & 45.50 & 49.93 \\
        \hline\hline
    \end{tabular}}
    \vspace{-10pt}
    \caption{PureNeurons ablation study on lesion extraction efficiency and lesion information delivery.}
    \label{tab:pure_neurons_ablation}
    \vspace{-15pt}
\end{table*}

\begin{figure}
    \centering
    \includegraphics[width=0.85\linewidth]{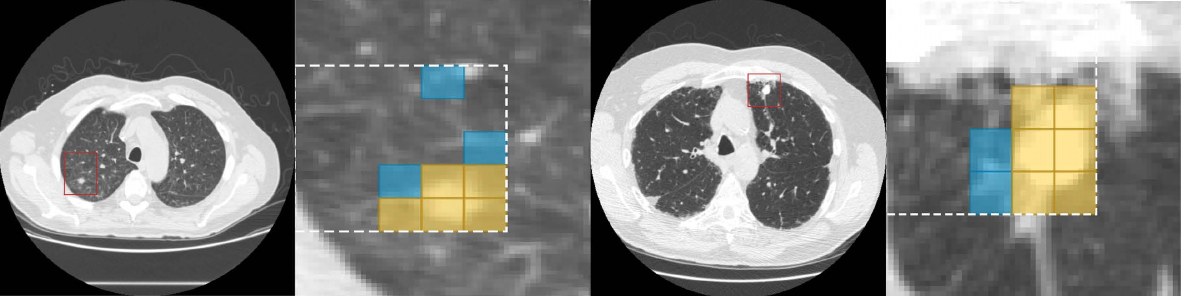}
    \vspace{-5pt}
    \caption{Visualization of lesion patch extraction by PureNeurons on LIDC-IDRI.}
    \label{fig:pure_neurons_visualization}
    \vspace{5pt}
    \centering
    \includegraphics[width=0.85\linewidth]{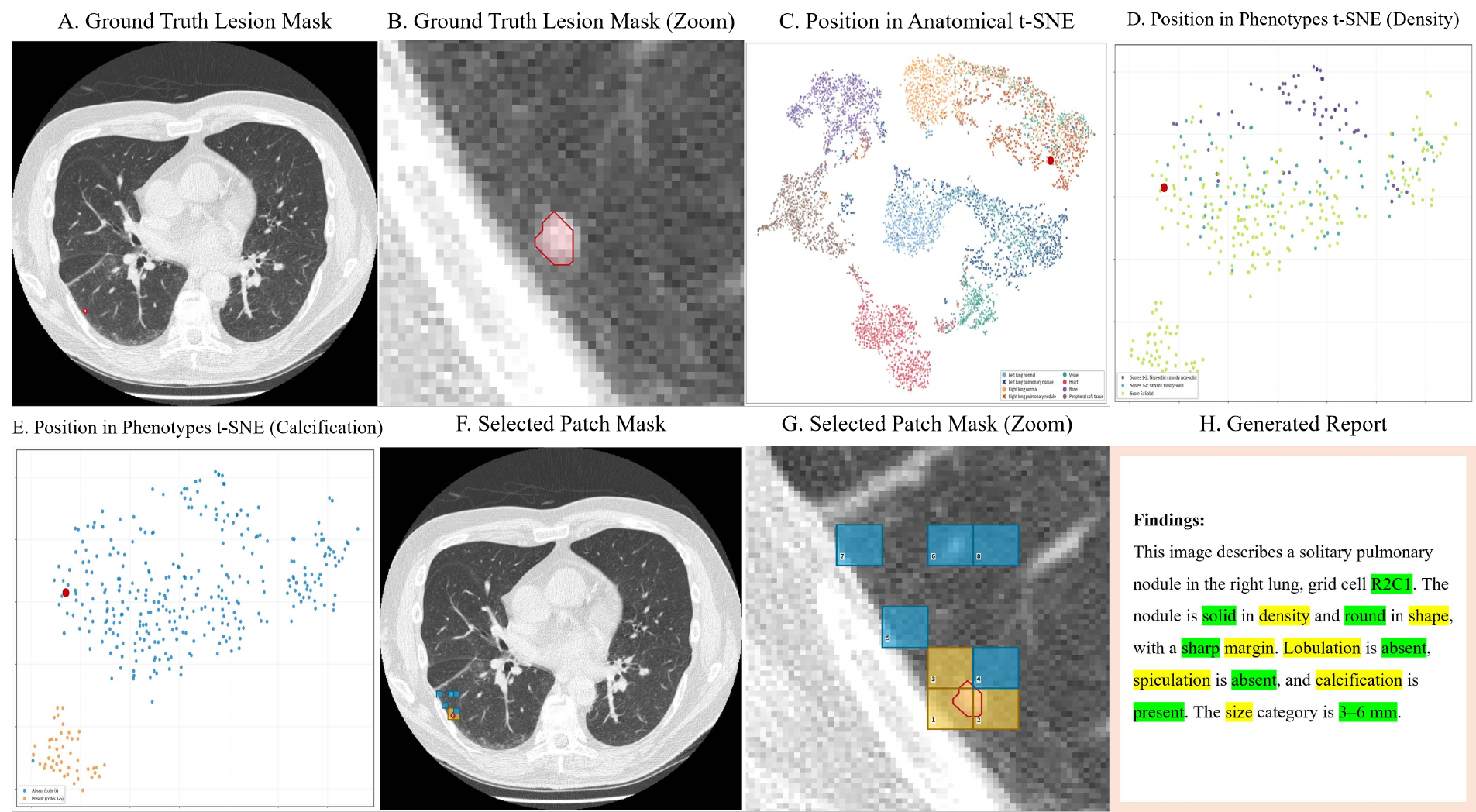}
    \vspace{-5pt}
    \caption{Case study of PureVision, illustrating the visual representations learned by PureEyes, lesion evidence selected by PureNeurons, and the resulting report generation.}
    \label{fig:case_study}
    \vspace{-15pt}
\end{figure}

\subsection{Lesion Extraction Efficiency and Information Delivery Ablation in PureNeurons}

Table~\ref{tab:pure_neurons_ablation} evaluates the lesion extraction efficiency and information delivery design of PureNeurons. Increasing the number of selected patches improves lesion patch recall from 60.11\% at $K=4$ to 68.84\% at $K=16$, while the redundant patch rate increases from 45.23\% to 81.23\%. Notably, the recall improvement becomes limited beyond $K=8$, indicating that a relatively small number of patches already captures most lesion-related evidence, whereas further increasing the selection budget mainly introduces redundant visual information. The information delivery ablation further shows consistent improvements from the original MedGemma1.5 to the different PureNeurons configurations. Grounding and phenotype accuracies increase from 11.50/10.50 to 32.50/36.57 without PureNeurons, further to 40.00/38.43 with lesion grounding, and finally to 45.50/49.93 with lesion grounding and soft semantic fusion. These results demonstrate the importance of both lesion-focused information extraction and soft phenotype fusion for effectively delivering localized pathological evidence to the language model. Representative qualitative cases illustrating the lesion regions selected by PureNeurons are shown in Figure~\ref{fig:pure_neurons_visualization}

\subsection{Case Study}
Figure~\ref{fig:case_study} presents an end-to-end case study of PureVision on LIDC-IDRI, illustrating the interaction between PureEyes, PureNeurons, and the downstream report generation task. PureEyes first organizes anatomical and pathological representations into structured latent visual spaces, where the projected embeddings show that the input sample is located within the corresponding anatomical and phenotype distributions. Based on these enhanced visual representations, PureNeurons identifies a compact set of lesion related patches that overlap with the target nodule region, providing localized visual evidence for subsequent reasoning. The selected lesion evidence is then delivered to the decoder together with the original image information, enabling the model to generate a report containing the correct lesion location and multiple pathological phenotype descriptions. This case demonstrates that PureVision can establish structured visual representations, selectively extract lesion relevant evidence, and effectively transfer this information to downstream tasks.

\section{Conclusion}
We introduced \textbf{PureVision}, a geometry-supervised visual representation learning framework for multi-phenotype lesion interpretation in medical VLMs. \textbf{PureEyes} structures anatomical and phenotypic representations with ideal geometries, while \textbf{PureNeurons} captures and delivers sparse lesion-specific evidence. Experiments across three datasets and multiple imaging modalities show consistent improvements in lesion grounding and phenotype characterization, demonstrating the effectiveness of geometry-aware visual representation learning for fine-grained medical image interpretation.

\subsection*{AI use statement}

In this work, we used generative AI tools for evaluating the report generation task. Specifically, GPT6-Astra was used as a structured parser to extract predicted lesion locations and phenotype labels from generated radiology reports. The extracted predictions were then compared with the reference annotations using the deterministic evaluation metrics described in Appendix~\ref{app:data}. GPT6-Astra was not used to assign performance scores directly. We did not use generative AI tools for generating synthetic datasets, developing theoretical models or conceptual frameworks, formulating mathematical claims, proposing or refining hypotheses, designing research methodology or experiments, implementing methods, assisting with translation, cleaning or reformatting datasets, or supporting qualitative and thematic data analysis. Proof-related uses are not applicable to this work. Additionally, we used generative AI tools for drafting parts of the research paper, improving readability, and suggesting the structure of the manuscript. We reviewed and revised all AI-assisted content, manually checked factual statements and citations against the original sources, and take full responsibility for the final content of this work, including text, claims, evaluations, and other artifacts produced with the aid of generative AI.

\subsection*{Ethics statement}

This work does not raise ethical concerns that require additional disclosure under the conference guidelines. It involves no new human-subject recruitment or collection of identifiable private information, and all datasets are used in accordance with their original access and usage requirements.

\subsection*{Reproducibility statement}

We provide an anonymous code repository at \url{https://anonymous.4open.science/r/purevision-06C2} to support reproduction of the methods and experiments presented in this work.

\bibliography{iclr2027_conference}

@article{chen2025large,
  title={Large language model with region-guided referring and grounding for CT report generation},
  author={Chen, Zhixuan and Bie, Yequan and Jin, Haibo and Chen, Hao},
  journal={IEEE transactions on Medical Imaging},
  volume={44},
  number={8},
  pages={3139--3150},
  year={2025},
  publisher={IEEE}
}

@inproceedings{xing2025mca,
  title={Mca-rg: Enhancing llms with medical concept alignment for radiology report generation},
  author={Xing, Qilong and Song, Zikai and Zhang, Youjia and Feng, Na and Yu, Junqing and Yang, Wei},
  booktitle={International Conference on Medical Image Computing and Computer-Assisted Intervention},
  pages={380--390},
  year={2025},
  organization={Springer}
}

@inproceedings{gu2025radalign,
  title={Radalign: Advancing radiology report generation with vision-language concept alignment},
  author={Gu, Difei and Gao, Yunhe and Zhou, Yang and Zhou, Mu and Metaxas, Dimitris},
  booktitle={International Conference on Medical Image Computing and Computer-Assisted Intervention},
  pages={484--494},
  year={2025},
  organization={Springer}
}

@inproceedings{huang2024refer,
  title={A refer-and-ground multimodal large language model for biomedicine},
  author={Huang, Xiaoshuang and Huang, Haifeng and Shen, Lingdong and Yang, Yehui and Shang, Fangxin and Liu, Junwei and Liu, Jia},
  booktitle={International Conference on Medical Image Computing and Computer-Assisted Intervention},
  pages={399--409},
  year={2024},
  organization={Springer}
}

@article{bai2024m3d,
  title={M3d: Advancing 3d medical image analysis with multi-modal large language models},
  author={Bai, Fan and Du, Yuxin and Huang, Tiejun and Meng, Max Q-H and Zhao, Bo},
  journal={arXiv preprint arXiv:2404.00578},
  year={2024}
}

@inproceedings{luo2025vividmed,
  title={Vividmed: Vision language model with versatile visual grounding for medicine},
  author={Luo, Lingxiao and Tang, Bingda and Chen, Xuanzhong and Han, Rong and Chen, Ting},
  booktitle={Proceedings of the 2025 Conference of the Nations of the Americas Chapter of the Association for Computational Linguistics: Human Language Technologies (Volume 1: Long Papers)},
  pages={1800--1821},
  year={2025}
}

@article{li2023llava,
  title={Llava-med: Training a large language-and-vision assistant for biomedicine in one day},
  author={Li, Chunyuan and Wong, Cliff and Zhang, Sheng and Usuyama, Naoto and Liu, Haotian and Yang, Jianwei and Naumann, Tristan and Poon, Hoifung and Gao, Jianfeng},
  journal={Advances in neural information processing systems},
  volume={36},
  pages={28541--28564},
  year={2023}
}

@article{zambrano2025clinically,
  title={A clinically accessible small multimodal radiology model and evaluation metric for chest X-ray findings},
  author={Zambrano Chaves, Juan Manuel and Huang, Shih-Cheng and Xu, Yanbo and Xu, Hanwen and Usuyama, Naoto and Zhang, Sheng and Wang, Fei and Xie, Yujia and Khademi, Mahmoud and Yang, Ziyi and others},
  journal={Nature Communications},
  volume={16},
  number={1},
  pages={3108},
  year={2025},
  publisher={Nature Publishing Group UK London}
}

@article{sellergren2025medgemma,
  title={Medgemma technical report},
  author={Sellergren, Andrew and Kazemzadeh, Sahar and Jaroensri, Tiam and Kiraly, Atilla and Traverse, Madeleine and Kohlberger, Timo and Xu, Shawn and Jamil, Fayaz and Hughes, C{\'\i}an and Lau, Charles and others},
  journal={arXiv preprint arXiv:2507.05201},
  year={2025}
}

@article{sellergren2026medgemma,
  title={Medgemma 1.5 technical report},
  author={Sellergren, Andrew and Gao, Chufan and Mahvar, Fereshteh and Kohlberger, Timo and Jamil, Fayaz and Traverse, Madeleine and Tono, Alberto and Sadjad, Bashir and Yang, Lin and Lau, Charles and others},
  journal={arXiv preprint arXiv:2604.05081},
  year={2026}
}

@article{fallahpour2025medrax,
  title={Medrax: Medical reasoning agent for chest x-ray},
  author={Fallahpour, Adibvafa and Ma, Jun and Munim, Alif and Lyu, Hongwei and Wang, Bo},
  journal={arXiv preprint arXiv:2502.02673},
  year={2025}
}

@inproceedings{liu2025argus,
  title={Argus: benchmarking and enhancing vision-language models for 3d radiology report generation},
  author={Liu, Che and Wan, Zhongwei and Wang, Yuqi and Shen, Hui and Wang, Haozhe and Zheng, Kangyu and Zhang, Mi and Arcucci, Rossella},
  booktitle={Findings of the Association for Computational Linguistics: ACL 2025},
  pages={16448--16460},
  year={2025}
}

@article{li2026aor,
  title={Aor: Anatomical ontology-guided reasoning for medical large multimodal model in chest x-ray interpretation},
  author={Li, Qingqiu and Cui, Zihang and Bae, Seongsu and Xu, Jilan and Yuan, Runtian and Zhang, Yuejie and Feng, Rui and Shen, Quanli and Zhang, Xiaobo and Gao, Shang and others},
  journal={Advances in Neural Information Processing Systems},
  volume={38},
  pages={19331--19371},
  year={2026}
}

@article{bannur2024maira,
  title={Maira-2: Grounded radiology report generation},
  author={Bannur, Shruthi and Bouzid, Kenza and Castro, Daniel C and Schwaighofer, Anton and Thieme, Anja and Bond-Taylor, Sam and Ilse, Maximilian and P{\'e}rez-Garc{\'\i}a, Fernando and Salvatelli, Valentina and Sharma, Harshita and others},
  journal={arXiv preprint arXiv:2406.04449},
  year={2024}
}

@article{kyung2025region,
  title={Region-Aware Multimodal Large Language Model via SlowFast Tokenization and Pseudo-Mask Guidance for 3D CT Report Generation},
  author={Kyung, Sunggu and Seo, Jinyoung and Lim, Hyunseok and Kim, Dongyeong and Park, Hyungbin and Sung, Jimin and Kim, Jihyun and Jo, Wooyoung and Nam, Yoojin and Kim, Namkug},
  journal={arXiv preprint arXiv:2506.23102},
  year={2025}
}

@article{armato2011lung,
  title={The lung image database consortium (LIDC) and image database resource initiative (IDRI): a completed reference database of lung nodules on CT scans},
  author={Armato III, Samuel G and McLennan, Geoffrey and Bidaut, Luc and McNitt-Gray, Michael F and Meyer, Charles R and Reeves, Anthony P and Zhao, Binsheng and Aberle, Denise R and Henschke, Claudia I and Hoffman, Eric A and others},
  journal={Medical physics},
  volume={38},
  number={2},
  pages={915--931},
  year={2011},
  publisher={Wiley Online Library}
}

@article{lee2017curated,
  title={A curated mammography data set for use in computer-aided detection and diagnosis research},
  author={Lee, Rebecca Sawyer and Gimenez, Francisco and Hoogi, Assaf and Miyake, Kanae Kawai and Gorovoy, Mia and Rubin, Daniel L},
  journal={Scientific data},
  volume={4},
  number={1},
  pages={170177},
  year={2017},
  publisher={Nature Publishing Group}
}

@inproceedings{sambara20253dreasonknee,
  title={3dreasonknee: Advancing grounded reasoning in medical vision language models},
  author={Sambara, Sraavya and Kim, Sung Eun and Zhang, Xiaoman and Luo, Luyang and Johri, Shreya and Baharoon, Mohammed and Ro, Du Hyun and Rajpurkar, Pranav},
  booktitle={Biocomputing 2026: Proceedings of the Pacific Symposium},
  pages={99--113},
  year={2025},
  organization={World Scientific}
}

@article{lester2012osteoarthritis,
  title={The osteoarthritis initiative: a NIH public--private partnership},
  author={Lester, Gayle},
  journal={HSS journal},
  volume={8},
  number={1},
  pages={62--63},
  year={2012},
  publisher={Springer}
}

@article{wu2025towards,
  title={Towards generalist foundation model for radiology by leveraging web-scale 2d\&3d medical data},
  author={Wu, Chaoyi and Zhang, Xiaoman and Zhang, Ya and Hui, Hui and Wang, Yanfeng and Xie, Weidi},
  journal={Nature Communications},
  volume={16},
  number={1},
  pages={7866},
  year={2025},
  publisher={Nature Publishing Group UK London}
}

@article{xu2026lingshu,
  title={Lingshu: Generalist Foundation Model for Unified Multimodal Medical Understanding and Reasoning},
  author={Xu, Weiwen and Chan, Hou Pong and Li, Long and Aljunied, Mahani and Yuan, Ruifeng and Wang, Jianyu and Xiao, Chenghao and Chen, Guizhen and Liu, Chaoqun and Li, Zhaodonghui and others},
  journal={IEEE Transactions on Pattern Analysis and Machine Intelligence},
  year={2026},
  publisher={IEEE}
}

@article{jiang2025hulu,
  title={Hulu-med: A transparent generalist model towards holistic medical vision-language understanding},
  author={Jiang, Songtao and Wang, Yuan and Song, Sibo and Hu, Tianxiang and Zhou, Chenyi and Pu, Bin and Zhang, Yan and Yang, Zhibo and Feng, Yang and Zhou, Joey Tianyi and others},
  journal={arXiv preprint arXiv:2510.08668},
  year={2025}
}
\bibliographystyle{iclr2027_conference}

\newpage
\appendix
\section{Benchmark and Evaluation}
\label{app:data}

\subsection{Benchmark Construction}

\begin{wrapfigure}{r}{0.52\linewidth}
    \centering
    \includegraphics[width=0.98\linewidth]{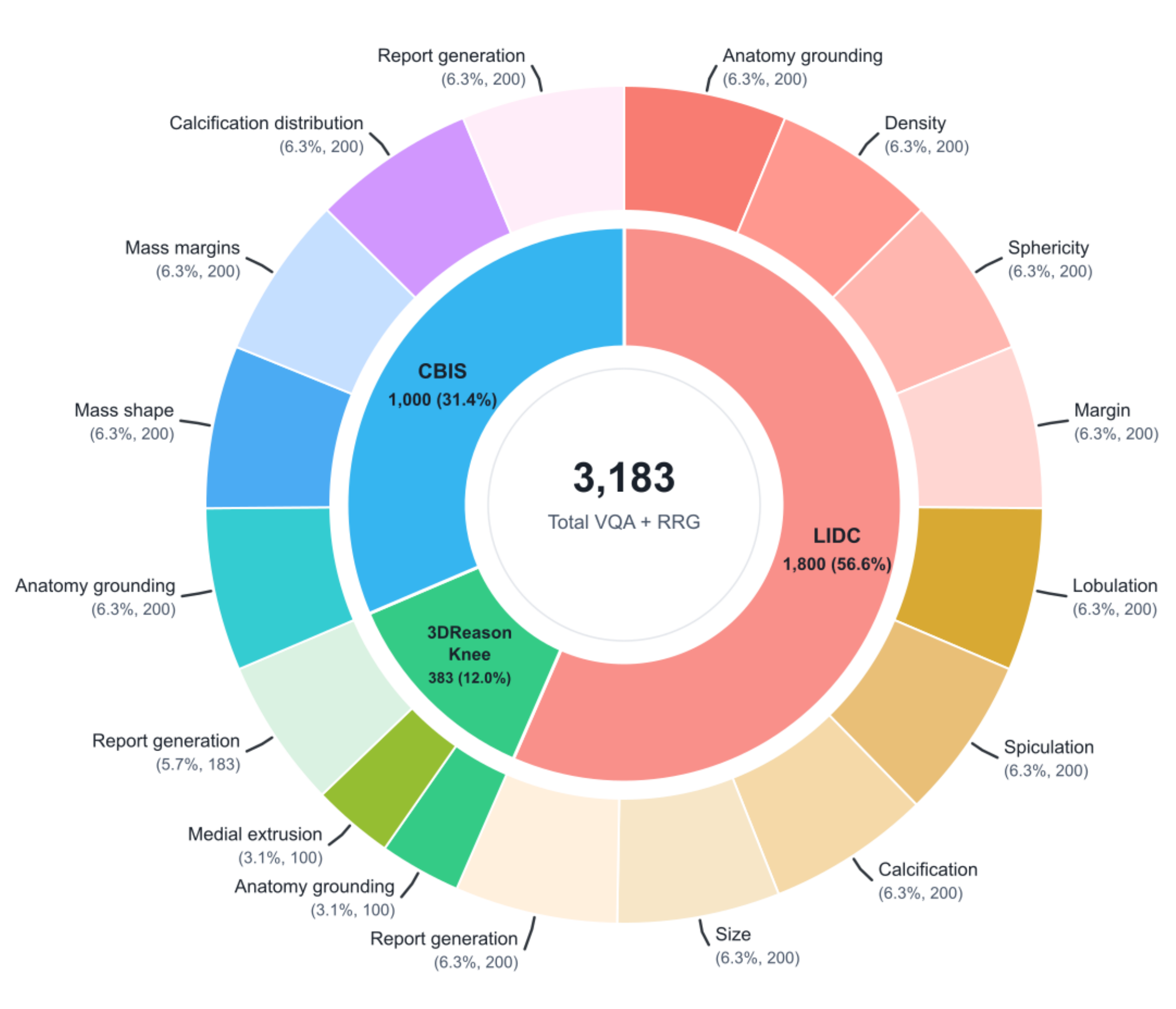}
    \caption{Distribution of benchmark instances by dataset and task.}
    \label{fig:benchmark_distribution}
\end{wrapfigure}

We construct VQA and RRG instances from patient-disjoint evaluation splits. 
For all three datasets, patients used for downstream evaluation are excluded from both representation learning and semantic alignment. 
Each evaluation image is associated with anatomical masks, a lesion mask, and phenotype labels, while the dataset-provided phenotype descriptions are directly used for RRG evaluation. 
Grounding questions use a $4\times4$ image grid, with ground-truth locations determined from the corresponding lesion masks. 
The benchmark distribution is shown in Fig.~\ref{fig:benchmark_distribution}.

For LIDC-IDRI, anatomical masks are generated using TotalSegmentator. 
We select single-nodule images whose nodule masks lie entirely within one grid cell and sample 200 grounding questions, each with one correct cell and three distractors. 
We additionally sample 200 questions for each of seven phenotypes: density, sphericity, margin, lobulation, spiculation, calcification, and size. 
Sampling is stratified by answer category within each dimension, yielding 1,600 VQA pairs.

For CBIS-DDSM, breast tissue masks are generated using GrabCut, pectoral muscle masks are generated by a pretrained Attention U-Net, and lesion masks are obtained from the original dataset annotations. 
We construct 200 grounding questions and 200 questions for each of three phenotypes: mass shape, mass margins, and calcification distribution. 
All VQA questions contain one correct answer and three distractors, yielding 800 VQA pairs.

For 3DReasonKnee, anatomical masks are obtained primarily from the dataset-provided nnU-Net segmentations. 
We construct 100 grounding questions and 100 questions assessing medial meniscus medial extrusion grades. 
All VQA questions contain one correct answer and three distractors, yielding 200 VQA pairs.

\subsection{Evaluation Metrics}

For VQA, predictions are matched against the reference answers. Grounding Accuracy is the proportion of correctly identified grid cells. Phenotypes Accuracy is computed by first calculating accuracy within each phenotype dimension and then averaging across dimensions:
\begin{equation}
    \operatorname{Acc}_{\mathrm{VQA}}=\frac{1}{G}\sum_{g=1}^{G}\frac{N_g^{\mathrm{correct}}}{N_g},
\end{equation}
where $G$ is 7 for LIDC-IDRI, 3 for CBIS-DDSM, and 1 for 3DReasonKnee. Missing or invalid answers are counted as incorrect.

For RRG, GPT6-Astra is used as a structured parser to extract the predicted grid location and phenotype labels from each generated report. The parser maps free-form report text to the predefined label space of each dataset. The extracted predictions are then evaluated deterministically against the reference annotations. We use a fixed evaluation prompt and deterministic decoding for all models and datasets. The parser is only allowed to return labels from the predefined candidate set, and missing or invalid outputs are counted as incorrect. Across all three datasets, Grounding Accuracy is the proportion of reports whose predicted grid cell exactly matches the reference cell in the $4\times4$ grid.

Phenotypes Accuracy is computed using the same formulation across all three datasets: balanced accuracy is calculated within each phenotype dimension and then averaged equally across dimensions:
\begin{equation}
    \operatorname{Acc}_{\mathrm{RRG}}=\frac{1}{G}\sum_{g=1}^{G}\frac{1}{|\mathcal{C}_g|}\sum_{c\in\mathcal{C}_g}\frac{N_{g,c}^{\mathrm{correct}}}{N_{g,c}},
\end{equation}
where $G$ is 7 for LIDC-IDRI, 3 for CBIS-DDSM, and 1 for 3DReasonKnee. $\mathcal{C}_g$ denotes the reference classes represented among applicable samples in dimension $g$, $N_{g,c}$ is the number of samples with reference class $c$, and $N_{g,c}^{\mathrm{correct}}$ counts predictions that exactly match this class. This formulation assigns equal weight to classes within each dimension and to the phenotype dimensions.

\section{Additional Experimental Results}
\label{app:additional_experiments}

\subsection{Detailed Phenotypes Accuracy}
\label{app:detailed_phenotypes_acc}

Tables~\ref{tab:lidc_phenotype_accuracy} and~\ref{tab:cbis_phenotype_accuracy} provide a detailed comparison of phenotype-specific VQA accuracy on LIDC-IDRI and CBIS-DDSM, respectively. PureVision improves most phenotype dimensions across the evaluated pretrained medical VLMs, indicating that the overall gains are broadly distributed across different pathological attributes rather than being dominated by a single phenotype. On LIDC-IDRI, when integrated with MedGemma1.5, PureVision increases the accuracies for density, margin, size, sphericity, calcification, lobulation, and spiculation from 25.50/16.00/5.00/5.00/8.00/10.00/4.00 to 34.50/27.00/45.00/29.00/83.50/60.50/70.00, with particularly pronounced gains in calcification, lobulation, and spiculation. Compared with existing representation alignment methods under the same MedGemma1.5 backbone, PureVision achieves the highest accuracy in five of the seven phenotype dimensions. On CBIS-DDSM, PureVision increases the mass shape, mass margins, and calcification distribution accuracies of MedGemma1.5 from 10.50/12.50/7.00 to 45.00/43.50/57.50 and achieves the highest accuracy across all three attributes among the compared alignment methods. Similar improvements across the remaining backbones further demonstrate the effectiveness of PureVision in fine-grained phenotype understanding across different diseases and imaging modalities. Since 3DReasonKnee contains only one phenotype dimension, namely medial meniscus medial extrusion, we do not provide an additional detailed phenotype-accuracy table for this dataset.

\begin{table}
  \centering
  \tiny
  \setlength{\tabcolsep}{1pt}
  \renewcommand{\arraystretch}{1.12}
  \resizebox{\textwidth}{!}{
  \begin{tabular}{@{\hspace{4pt}}l|c|c|c|c|c|c|c@{\hspace{4pt}}}
    \hline\hline
    \textbf{Models (VQA)} & \textbf{Density} & \textbf{Margin} & \textbf{Size} & \textbf{Sphericity} & \textbf{Calcification} & \textbf{Lobulation} & \textbf{Spiculation} \\ \hline
    RadFM & 4.00 & 3.00 & 3.00 & 5.50 & 7.00 & 8.50 & 4.00 \\
    \rowcolor{PVYellow} \quad \textbf{w/ PureVision} & \pv{11.00}{7.00} & \pv{5.00}{2.00} & \pv{5.00}{2.00} & \pv{33.00}{27.50} & \pv{40.00}{33.00} & \pv{50.50}{42.00} & \pv{16.50}{12.50} \\ \hline
    LLaVA-Med & 17.50 & 7.50 & 5.00 & 39.00 & 46.50 & 56.50 & 22.50 \\
    \rowcolor{PVYellow} \quad \textbf{w/ PureVision} & \pv{20.50}{3.00} & \pv{42.50}{35.00} & \pv{27.00}{22.00} & \pv{83.00}{44.00} & \pv{60.50}{14.00} & \pv{60.00}{3.50} & \pvloss{20.50}{2.00} \\ \hline
    Lingshu & 23.50 & 13.50 & 10.00 & 45.50 & 52.50 & 63.00 & 29.00 \\
    \rowcolor{PVYellow} \quad \textbf{w/ PureVision} & \pv{42.50}{19.00} & \pv{27.50}{14.00} & \pv{44.50}{34.50} & \pvloss{30.50}{15.00} & \pv{80.00}{27.50} & \pvloss{60.00}{3.00} & \pv{69.00}{40.00} \\ \hline
    Hulu-Med & 32.00 & 20.50 & 34.50 & 25.00 & 73.50 & 50.00 & 59.50 \\
    \rowcolor{PVYellow} \quad \textbf{w/ PureVision} & \pv{42.50}{10.50} & \pv{27.00}{6.50} & \pv{44.00}{9.50} & \pv{30.00}{5.00} & \pv{84.00}{10.50} & \pv{60.50}{10.50} & \pv{70.00}{10.50} \\ \hline
    MedGemma & 32.00 & 19.00 & 33.50 & 25.00 & 22.50 & 61.50 & 50.00 \\
    \rowcolor{PVYellow} \quad \textbf{w/ PureVision} & \pv{42.50}{10.50} & \pv{28.00}{9.00} & \pv{44.00}{10.50} & \pv{31.50}{6.50} & \pv{84.00}{61.50} & \pvloss{60.50}{1.00} & \pv{70.00}{20.00} \\ \hline
    MedGemma1.5 & 25.50 & 16.00 & 5.00 & 5.00 & 8.00 & 10.00 & 4.00 \\
    \quad w/ RadAlign & 22.50 & 12.50 & 9.00 & 44.50 & 52.00 & 62.00 & 28.00 \\
    \quad w/ MCA-RG & 27.50 & 17.50 & 31.00 & 25.00 & 23.00 & 43.00 & 38.00 \\
    \quad w/ Reg2RG & 33.00 & 9.50 & 0.00 & 25.00 & 12.50 & 50.00 & 36.50 \\
    \rowcolor{PVYellow} \quad \textbf{w/ PureVision} & \pv{34.50}{9.00} & \pv{27.00}{11.00} & \pv{45.00}{40.00} & \pv{29.00}{24.00} & \pv{83.50}{75.50} & \pv{60.50}{50.50} & \pv{70.00}{66.00} \\ \hline\hline
  \end{tabular}}
  \caption{Detailed pulmonary nodule phenotype accuracy (\%) on LIDC-IDRI.}
  \label{tab:lidc_phenotype_accuracy}
\end{table}

\begin{table}
  \centering
  \tiny
  \setlength{\tabcolsep}{4pt}
  \renewcommand{\arraystretch}{1.18}
  \resizebox{0.82\textwidth}{!}{
  \begin{tabular}{@{\hspace{4pt}}l|c|c|c@{\hspace{4pt}}}
    \hline\hline
    \textbf{Models (VQA)} & \textbf{Mass Shape} & \textbf{Mass Margins} & \textbf{Calcification Distribution} \\ \hline
    RadFM & 11.00 & 18.50 & 21.50 \\
    \rowcolor{PVYellow} \quad \textbf{w/ PureVision} & \pv{33.50}{22.50} & \pvloss{12.50}{6.00} & \pv{26.00}{4.50} \\ \hline
    LLaVA-Med & 33.50 & 27.50 & 31.50 \\
    \rowcolor{PVYellow} \quad \textbf{w/ PureVision} & \pvloss{33.00}{0.50} & \pv{32.00}{4.50} & \pv{47.50}{16.00} \\ \hline
    Lingshu & 19.00 & 4.00 & 13.00 \\
    \rowcolor{PVYellow} \quad \textbf{w/ PureVision} & \pv{35.50}{16.50} & \pv{18.50}{14.50} & \pv{52.50}{39.50} \\ \hline
    Hulu-Med & 13.00 & 3.50 & 6.50 \\
    \rowcolor{PVYellow} \quad \textbf{w/ PureVision} & \pv{31.50}{18.50} & \pv{22.50}{19.00} & \pv{52.00}{45.50} \\ \hline
    MedGemma & 10.00 & 12.00 & 3.50 \\
    \rowcolor{PVYellow} \quad \textbf{w/ PureVision} & \pv{24.50}{14.50} & \pv{20.50}{8.50} & \pv{51.00}{47.50} \\ \hline
    MedGemma1.5 & 10.50 & 12.50 & 7.00 \\
    \quad w/ RadAlign & 23.50 & 20.50 & 49.50 \\
    \quad w/ MCA-RG & 14.00 & 16.50 & 28.50 \\
    \quad w/ Reg2RG & 12.50 & 24.50 & 14.50 \\
    \rowcolor{PVYellow} \quad \textbf{w/ PureVision} & \pv{45.00}{34.50} & \pv{43.50}{31.00} & \pv{57.50}{50.50} \\ \hline\hline
  \end{tabular}}
  \caption{Detailed breast lesion phenotype accuracy (\%) on CBIS-DDSM.}
  \label{tab:cbis_phenotype_accuracy}
\end{table}

\subsection{Fine-Grained Distribution of Latent Vision Representation of Pulmonary Patch Representations}
\label{app:appendix_pulmonary_tsne}

\begin{figure}[H]
    \centering
    \includegraphics[width=0.8\linewidth]{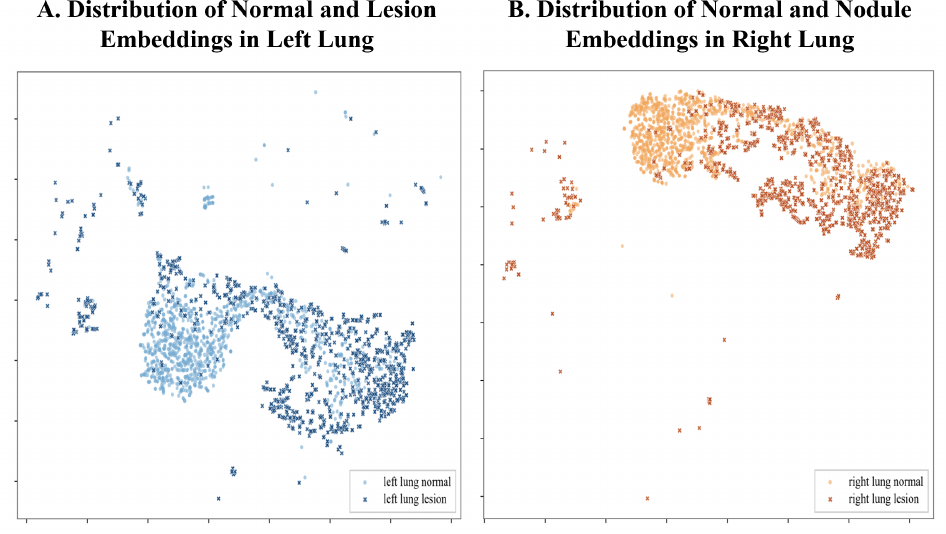}
    \caption{Fine-Grained Separation of Normal and Lesion Lung Representations}
    \label{fig:finegrained_separation_1}
    \vspace{-15pt}
\end{figure}

\begin{figure}[H]
    \centering
    \includegraphics[width=0.8\linewidth]{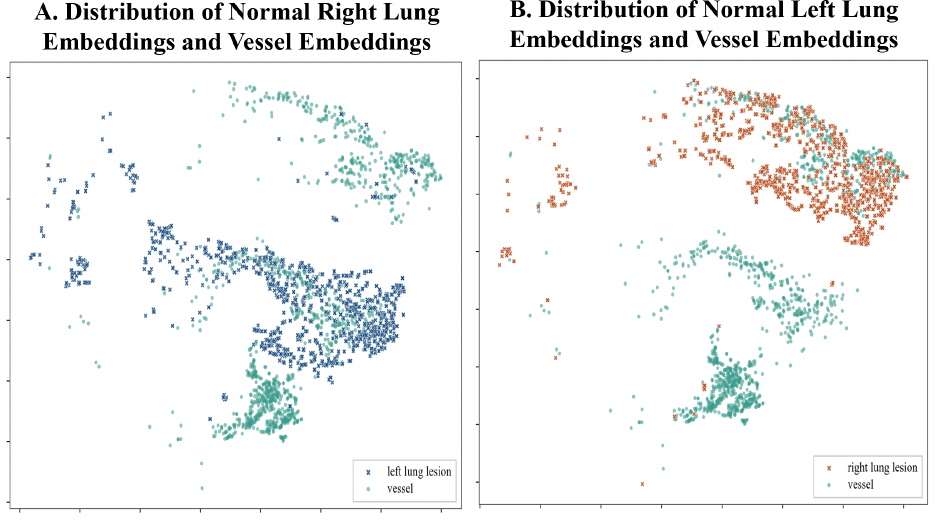}
    \caption{Fine-Grained Organization of Pulmonary Lesion and Vessel Representations}
    \label{fig:finegrained_separation_2}
    \vspace{-15pt}
\end{figure}

To further examine the fine-grained organization of the visual space learned by PureEyes, we isolate several pulmonary patch categories from the LIDC-IDRI t-SNE projection. As shown in Figures~\ref{fig:finegrained_separation_1} and~\ref{fig:finegrained_separation_2}, normal and abnormal lung patches form distinguishable subregions within the broader left- and right-lung clusters. In both lungs, most normal patches are concentrated in relatively compact regions, whereas lesion patches occupy different portions of the embedding space, demonstrating that PureEyes can capture pathological variations while preserving their anatomical context. By contrast, pulmonary lesion and vessel patches exhibit partial overlap, which is expected because nodules and cross-sectional vessels may have similar local appearances within a single CT slice. Nevertheless, a substantial proportion of vessel patches form vessel-dominated subclusters that are clearly separated from lesion patches, particularly in the peripheral regions of the embedding space. This observation suggests that the representations learned by PureEyes incorporate not only the local visual appearance of each patch but also contextual information from its surrounding anatomy. Together, these visualizations provide additional qualitative evidence that PureEyes organizes pulmonary patches according to both anatomical identity and fine-grained pathological characteristics.

\subsection{Additional Phenotypes Distribution}
\label{app:additional_phenotype_distribution}

\begin{figure}
    \centering
    \includegraphics[width=\linewidth]{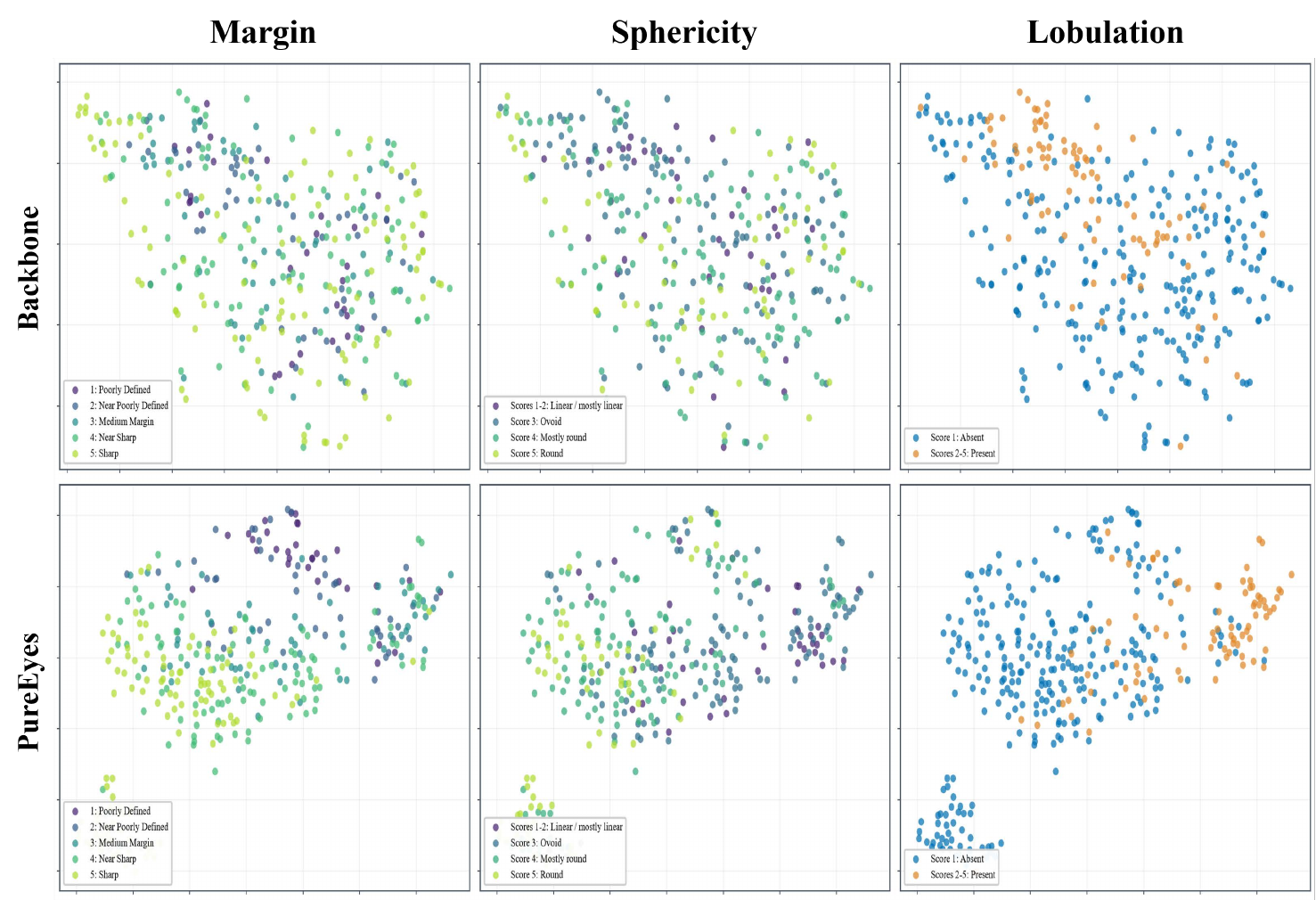}
    \caption{Additional t-SNE visualizations of phenotype representations on LIDC-IDRI.}
    \label{fig:addition_phenotype_distribution}
\end{figure}

Figure~\ref{fig:addition_phenotype_distribution} presents additional t-SNE visualizations for the remaining LIDC-IDRI phenotypes, including sphericity, margin, and lobulation. With the original MedGemma1.5 vision encoder, samples assigned different phenotype scores are substantially intermixed, and the latent distributions exhibit little correspondence with the underlying phenotype levels. In contrast, the representations learned with PureEyes show markedly clearer organization. Samples with similar phenotype scores tend to remain close in the embedding space, whereas samples exhibiting larger differences in sphericity, margin, or lobulation are distributed into more distinguishable regions. In particular, the learned distributions preserve gradual transitions between adjacent phenotype levels while increasing the separation between visually distinct manifestations. These results are consistent with the examples reported in the main text and further demonstrate that PureEyes learns a structured phenotype representation space across all evaluated pulmonary nodule characteristics, rather than improving only a small subset of attributes.

\section{Implementation Details}
\label{app:implementation}

\subsection{Architecture and training stages}
For every medical VLM backbone, the anatomical encoder $E_A$ and phenotypic encoder $E_P$ are initialized independently from the backbone vision encoder and optimized as separate visual encoders. For the MedGemma-1.5 implementation, both encoders use the complete SigLIP vision tower with $896\times896$ inputs, patch size 14, a $64\times64$ patch grid, and 1,152-dimensional visual tokens. PureEyes updates the complete visual towers during representation learning. Afterward, both encoders are frozen, and a shared alignment layer consisting of RMS normalization followed by a bias-free $1{,}152\!\rightarrow\!2{,}560$ projection is trained to map anatomical and phenotypic patch representations into the language-compatible space. No LoRA or additional residual adapter is used. Training masks are used only to identify supervised patches and are never provided during inference.

\subsection{Optimization}
All stages use AdamW with weight decay 0.01, 5\% linear warm-up followed by cosine decay, gradient clipping at 1.0, and BF16 mixed precision. Phenotypic representation learning uses a learning rate of $5\times10^{-7}$ for 30 epochs with a global batch size of 16. Anatomical representation learning uses a learning rate of $2\times10^{-6}$, and the semantic alignment stage uses a learning rate of $2\times10^{-5}$ for 30 epochs. During semantic alignment, the visual encoders and language model remain frozen, and only the shared RMS normalization and projection parameters are optimized.

\subsection{Distance definition and robust penalty}
All phenotype geometry objectives operate on $\ell_2$-normalized representations,
\begin{equation}
    \bar{\mathbf e}_i^g =
    \frac{\mathbf e_i^g}{\|\mathbf e_i^g\|_2},
    \qquad
    d_{ij}^g =
    \left\|
    \bar{\mathbf e}_i^g-
    \bar{\mathbf e}_j^g
    \right\|_2 ,
\end{equation}
where $g$ denotes a phenotype dimension. We use the Smooth-$L_1$ penalty
\begin{equation}
\rho_{\delta}(r)=
\begin{cases}
\dfrac{r^2}{2\delta}, & |r|<\delta,\\[4pt]
|r|-\dfrac{\delta}{2}, & \text{otherwise},
\end{cases}
\end{equation}
with $\delta=1.0$ throughout. For normalized embeddings, this Euclidean distance satisfies
$d_{ij}^2=2-2\cos(\bar{\mathbf e}_i^g,\bar{\mathbf e}_j^g)$ and therefore preserves the angular geometry of the pretrained representation while providing a bounded distance space for explicit geometric supervision.

\subsection{Continuous phenotype geometry}
For phenotypes associated with continuous physical measurements, we supervise visual distances using the corresponding measurement differences. In LIDC-IDRI, lesion size is treated as a continuous phenotype and is represented by its physical diameter in millimeters. The measurement is normalized using statistics computed from the training split,
\begin{equation}
    \tilde y_i^g =
    \frac{y_i^g-y_{\min}^g}
    {y_{\max}^g-y_{\min}^g},
\end{equation}
and the target distance between two samples is defined as
\begin{equation}
    d_{ij}^{g,*}
    =
    \alpha_g
    \left|
    \tilde y_i^g-\tilde y_j^g
    \right|,
\end{equation}
where $\alpha_g=1.5$ matches the maximum target distance used by the discrete phenotype objectives. The continuous geometry loss is
\begin{equation}
    \mathcal L_{\mathrm{cont}}^g
    =
    \frac{1}{|\mathcal P_g|}
    \sum_{(i,j)\in\mathcal P_g}
    \rho_{\delta}
    \left(
        d_{ij}^g-d_{ij}^{g,*}
    \right).
\end{equation}
This objective preserves metric information in the phenotype space: lesions with similar physical measurements are encouraged to remain nearby, while larger measurement differences correspond to proportionally larger visual distances.

\subsection{Ordinal phenotype geometry}
For ordinal phenotypes, including density, sphericity, margin, lobulation, and spiculation in LIDC-IDRI and medial meniscus extrusion grade in 3DReasonKnee, the target pairwise distance is determined by the separation between their ordered labels. For an attribute with $C_g$ levels,
\begin{equation}
    d_{ij}^{g,*}
    =
    1.5
    \frac{|y_i^g-y_j^g|}
    {C_g-1},
\end{equation}
and the pairwise objective is
\begin{equation}
    \mathcal L_{\mathrm{pair}}^g
    =
    \frac{1}{|\mathcal P_g|}
    \sum_{(i,j)\in\mathcal P_g}
    \rho_{\delta}
    \left(
        d_{ij}^g-d_{ij}^{g,*}
    \right).
\end{equation}
We further impose an ordinal ranking constraint. For anchor $i$, let
$\bar d_{i,c}$ denote its mean distance to samples belonging to category $c$.
For two categories $c_{\mathrm{near}}$ and $c_{\mathrm{far}}$ satisfying
$\Delta_{\mathrm{near}}<\Delta_{\mathrm{far}}$, where
$\Delta_c=|y_i^g-c|$, we define
\begin{equation}
\mathcal L_{\mathrm{rank}}^g
=
\frac{1}{|\mathcal R_g|}
\sum
\left[
\bar d_{i,c_{\mathrm{near}}}
-
\bar d_{i,c_{\mathrm{far}}}
+
m_o
\left(
\Delta_{\mathrm{far}}-\Delta_{\mathrm{near}}
\right)
\right]_+ ,
\end{equation}
with $m_o=0.1$. The final ordinal objective is
\begin{equation}
    \mathcal L_{\mathrm{ord}}^g
    =
    \mathcal L_{\mathrm{pair}}^g
    +
    \mathcal L_{\mathrm{rank}}^g .
\end{equation}

\subsection{Categorical phenotype geometry}
For categorical phenotypes without an ordinal relationship, the target distance is zero for samples from the same category and 1.5 for samples from different categories:
\begin{equation}
d_{ij}^{g,*}
=
\begin{cases}
0, & y_i^g=y_j^g,\\
1.5, & y_i^g\neq y_j^g.
\end{cases}
\end{equation}
The corresponding objective is
\begin{equation}
    \mathcal L_{\mathrm{cat}}^g
    =
    \frac{1}{|\mathcal P_g|}
    \sum_{(i,j)\in\mathcal P_g}
    w_{ij}^g
    \rho_{\delta}
    \left(
        d_{ij}^g-d_{ij}^{g,*}
    \right),
\end{equation}
where $w_{ij}^g$ compensates for category imbalance. Calcification is treated as categorical in LIDC-IDRI. For CBIS-DDSM, mass shape, mass margins, and calcification distribution are treated as categorical phenotypes.

For all phenotype objectives, each anchor is paired with samples from the current batch together with a memory bank containing the most recent 1,024 detached lesion representations. Self-pairs are excluded. The seven LIDC-IDRI phenotype dimensions use normalized weights of $0.125$, $0.125$, $0.125$, $0.25$, $0.1875$, $0.09375$, and $0.09375$ for margin, calcification, density, sphericity, size, lobulation, and spiculation, respectively.

\subsection{Anatomical geometry objectives}
The anatomical encoder is optimized with
\begin{equation}
\mathcal L_A
=
4\mathcal L_{\mathrm{anatomy}}
+
0.55\mathcal L_{\mathrm{parent}}
+
4\mathcal L_{\mathrm{hierarchy}}
+
10\mathcal L_{\mathrm{distill}}
+
\mathcal L_{\mathrm{conf}} .
\end{equation}
$\mathcal L_{\mathrm{anatomy}}$ organizes the anatomical categories according to a regular-simplex target geometry, while $\mathcal L_{\mathrm{parent}}$ preserves the geometry of their higher-level anatomical parent regions. $\mathcal L_{\mathrm{hierarchy}}$ further separates normal and lesion states within the same anatomical parent. To avoid unnecessary deformation of normal anatomical representations, $\mathcal L_{\mathrm{distill}}$ minimizes the cosine distance between the learned encoder and a frozen pretrained teacher on non-lesion patches.

\subsection{Confounding-tissue separation}
Some normal anatomical structures exhibit local appearances similar to lesions and can therefore occupy neighboring regions in the pretrained representation space. We explicitly increase their separation using a teacher-relative confounding-tissue objective. Let
$s_{ij}^{A}$ and $s_{ij}^{T}$ denote the cosine similarities between a target anatomical patch and a confounding-tissue patch under the learned anatomical encoder and frozen teacher, respectively. We define
\begin{equation}
\mathcal L_{\mathrm{sep}}
=
\frac{1}{|\mathcal Q_{\mathrm{conf}}|}
\sum_{(i,j)\in\mathcal Q_{\mathrm{conf}}}
\left[
s_{ij}^{A}
-
s_{ij}^{T}
+
m_c
\right]_+ ,
\end{equation}
where $m_c=0.01$. Thus, the learned representation is required to reduce the similarity of a confounding pair by at least $m_c$ relative to the pretrained teacher, rather than forcing unrelated anatomical representations to move arbitrarily far apart.

We additionally apply supervised contrastive regularization to improve the compactness of the confounding anatomical categories:
\begin{equation}
    \mathcal L_{\mathrm{conf}}
    =
    \mathcal L_{\mathrm{sep}}
    +
    0.05\,
    \mathcal L_{\mathrm{SupCon}},
\end{equation}
with temperature $0.1$. For LIDC-IDRI, pulmonary vessels are treated as the primary confounding tissue for left- and right-lung regions because vessel cross-sections can exhibit local appearances similar to pulmonary nodules.

\subsection{Semantic alignment and lesion selection}
After PureEyes training, both visual encoders are frozen. Anatomical and phenotypic patch representations share the same alignment layer. Text targets are encoded by the frozen language model, and the aligner is trained using temperature-scaled cosine classification with temperature $0.07$ together with a cosine attraction term to the corresponding text representation. At inference time, the anatomical correspondence between patch $i$ and label $c$ is
\begin{equation}
S_{i,c}^{A}
=
\operatorname{cos}
\left(
G(\mathbf h_i^A),\,
\mathbf t_c^A
\right).
\end{equation}
The lesion score is
\begin{equation}
\rho_i
=
\max_{c\in\mathcal C_{\mathrm{lesion}}^A}S_{i,c}^{A}
-
\max_{c\in\mathcal C_{\mathrm{other}}^A}S_{i,c}^{A}.
\end{equation}
PureNeurons first identifies the most lesion-consistent local $5\times5$ neighborhood and then selects the top $K=8$ patches within this region. Anatomical and phenotypic representations are matched using their shared spatial patch indices.

\subsection{Soft semantic fusion}
For phenotype group $g$, similarities over the selected patches are first aggregated as
\begin{equation}
s_{g,c}
=
\sum_{i\in\mathcal R}
w_iS_{i,c}^{g},
\end{equation}
where $w_i$ is obtained from a softmax over the selected lesion scores. The aggregated similarities are standardized within each phenotype group and converted into subtype probabilities:
\begin{equation}
p_{g,c}
=
\operatorname{softmax}_{c}
\left(
\frac{
(s_{g,c}-\mu_g)/\sigma_g
}{
\tau_s
}
\right),
\qquad
\tau_s=0.125.
\end{equation}
The soft semantic representation is then constructed as
\begin{equation}
\mathbf v_{g,\ell}
=
\sum_{c\in\mathcal C_g}
p_{g,c}\,
\bar{\mathbf u}_{g,c,\ell},
\end{equation}
where shorter candidate sequences are extended by repeating their final token embedding to obtain a common length within each phenotype group. The fused anatomical and phenotype tokens are inserted after the visual tokens and before the task instruction.

\subsection{Computational Resources}

All experiments were conducted on NVIDIA RTX PRO 6000 Blackwell Server Edition GPUs. Phenotypic representation learning used two GPUs with distributed data parallel training, while anatomical representation learning and semantic alignment were each performed on a single GPU. Downstream inference and evaluation used two GPUs. All stages used BF16 mixed precision where applicable.

\end{document}